\documentclass[letterpaper, 10 pt, conference]{ieeeconf}  

\IEEEoverridecommandlockouts                              

\usepackage{graphicx}

\usepackage[skins]{tcolorbox}

\title{\LARGE \bf
Towards Abstract Relational Learning in Human Robot Interaction
}

\author{Mohamadreza Faridghasemnia$^{1}$, Daniele Nardi$^{2}$ and Alessandro Saffiotti$^{1}$
\thanks{This work was partially supported by the Wallenberg AI, Autonomous Systems and Software Program (WASP) funded
by the Knut and Alice Wallenberg Foundation.}
\thanks{$^{1}$Center for Applied Autonomous Sensor Systems (AASS), Örebro University, Örebro, Sweden
        {\tt\small \{mohamadreza.farid, alessandro.saffiotti\}@oru.se}}%
\thanks{$^{2}$Department of Computer, Control and Management Engineering “Antonio
Ruberti”, Sapienza University of Rome, Rome, Italy
        {\tt\small nardi@dis.uniroma1.it}}%
}

\begin{document}

\maketitle
\thispagestyle{empty}
\pagestyle{empty}

\begin{abstract}
Humans have a rich representation of the entities in their environment. Entities are described by their attributes, and entities that share attributes are often semantically related.  For example, if two books have ``Natural Language Processing'' as value of their `title' attribute, we can expect that their `topic' attribute will also be equal, namely, ``NLP''.  Humans tend to generalize such observations, and infer sufficient conditions under which the `topic' attribute of any entity is ``NLP''.  If robots need to interact successfully with humans, they need to represent entities, attributes, and generalizations in a similar way. This ends in a contextualized cognitive agent that can adapt its understanding, where context provides sufficient conditions for a correct understanding. In this work, we address the problem of how to obtain these representations through human-robot interaction.  We integrate visual perception and natural language input to incrementally build a semantic model of the world, and then use inductive reasoning to infer logical rules that capture generic semantic relations, true in this model.  These relations can be used to enrich the human-robot interaction, to populate a knowledge base with inferred facts, or to remove uncertainty in the robot's sensory inputs.
\end{abstract}

\section{Introduction}

As the growth of robots interacting with humans, different levels of environment understanding is required by the robot. A robot acting in an environment has to deal with many open questions, thus needs different levels of reasoning to do a task. Usually, robots rely on their initial knowledge, perception and their cognitive abilities to be able to understand and do reasoning in their situated environment. A recently hooked topic to a better Knowledge-Based cognition is dialogic interaction between a human and a robot, where the robot captures fresh information about the environment from a user through Natural Language. Information comes from Natural Language together with visually perceived information, and a Knowledge Base (KB) lets a cognitive agent reach different levels of understanding in the environment. 

The first level of understanding can be seen as classification and detection on sensory inputs, e.g. detection of objects in visual perception, or role tagging of lexical in a sentence. The second level of understanding concerns finding relations between different sensory inputs, e.g. finding common attributes in language and vision. Some famous problems such as symbol grounding \cite{harnad1990symbol} and anchoring \cite{coradeschi2003introduction} concern finding correspondences in different sensory input modalities. A higher and abstract level of understanding can be thought to find relations between the entities in an environment. e.g. in a scene with a desk and a book on top, some of the relationships between these are their relative physical position and their semantics that shows how entities (book and desk) are similar. 

Understanding relationships between physical entities can also be extended to the attributes of entities. Indeed the same definition of the relationship between entities can be found for the attributes. For example, when a user declares \textit{freshness} attribute of 'apple-1' is 'spoiled'(the value of \textit{freshness} attribute), as well as 'apple-2' and 'apple-3', but 'orange-1' and 'banana-1' are 'fresh', a relation between the values of \textit{freshness} attribute exists which connects semantic of entities; In this example, is that all apples are 'spoiled', and the rest of fruits are 'fresh', with closed world assumption. (In this paper \textit{attributes} are in italic font, and 'attribute values' are in quotation marks.)

Relation and rules for attributes of entities can help a robot that is interacting with a human in many applications. For example when a user utters "bring me a fruit", using the rules obtained for \textit{freshness} attribute, the robot notices which fruits are spoiled and which are fresh to eat. Such logical rules between attributes let the robot realize that apples are spoiled, apples should be thrown out, and added to the shopping list. Moreover, the obtained rule for attributes can be used in a robot's low-level sensory input processing. Consider an utterance where the user of our example is declaring that a physical entity is spoiled, but the robot's visual perception has doubt whether the perceived object is apple or pear. As the robot already found that all apples are spoiled and other fruits are fresh, so the perceptual detection refines the recognized object as the apple.


In this work, we propose a framework for learning logical rules that represent relations between attributes in a semantic model of the robot's environment. Such logical rules help the robot to find which attributes (e.g. properties of objects in a scene) entail a specific attribute. A distinctive novelty of our work is to generalize rules from a semantic model built via Human-Robot Interaction (HRI), through the integration of visual and linguistic cues. Our framework goes all the way from sensory input data to abstract First-Order Logic formulas that describe abstract relationship between attributes of entities in a scene. 
We focus on \emph{latent rules}, which the robot can capture implicitly when a human describes objects to the robot. In other words, we do not require the user to give rules explicitly to the robot, but rather we let the robot find rules and do further reasoning based on self-computed rules for improving its interaction with the user.  

This paper continues with the review of related work, and then in Section \ref{sec:schema} the proposed framework is described, followed by an implementation to demonstrate the viability of the proposed framework in Section \ref{sec:proposedFramework}. In Section \ref{sec:demo} results of a test scenario are given, followed by the discussion about the applicability of the framework. In the end, conclusions of this work are drawn.

\section{Related Work}
\label{sec:relatedWork}
Our framework combines insights and techniques from semantic scene understanding; from situated Natural Language dialogue; and from Inductive Logic. Accordingly, in this section we review some of the related works. Firstly, we review some works around understanding a scene alongside a sentence. Thereafter, we review works in the joint fields of Natural Language and computer vision, with the focus on their Natural Language understanding module. Lastly, we discuss works that concern reasoning and Inductive Logic. 

As Neural Network advancements achieved astonishing results, a new trend started to understand a scene via objects and their attributes in images. This trend aims to name and detect objects, and describe the attributes and their relationship in a scene \cite{krishnavisualgenome}. The work described in \cite{Isola2015DiscoveringSA} focus on the state and transformation of objects in a scene for understanding an image. Sadeghi et al. in \cite{Sadeghi2011Recognition} point to depicted interactions for understanding an image. Some works, for example the work described in \cite{OlgaAttribute2012}, solely focus on detection of the attributes from image, including learning visual relations between objects in the image. Alongside this trend, researchers advised to shift the task of object recognition by names to recognition by descriptions \cite{Farhadi2009Describing}; which transforms the problems of finding attributes and symbol grounding to the problem of Referential Expression Grounding (REG). Some of the most promising works in this field are  \cite{Kazemzadeh2014Refer}, \cite{Mao2016Generation}, \cite{Yu2018Mattnet}, which try to find the referred object, given a phrase that describes object via attributes and interactions showed in the image. Despite the fact that all of these works do their task based on attributes of objects, they treat them as visual features for describing depicted objects, which bound their domain of attributes to visually perceivable attributes; In this work we extend the domain of attributes by including attributes captured from dialog.

Linguistics have different points of view to attributes. Some works extract the logical form of a sentence, through semantic parser \cite{kushman2013using},  \cite{liang2016learning}, \cite{DongLanguage2016}, \cite{steedman2011combinatory}. The logical form that is computed from a language may be used for obtaining predicates from language, or obtaining attributes of a symbol in a sentence. 
Some works use the combination of Natural Language and computer vision. The work presented in \cite{vanzo2018incrementally} captures semantic attributes from Natural Language, focused on category of objects inferred from a visual classifier. Pronobis et al. in \cite{pronobis2012large} use attributes for resolving ambiguity in semantic mapping, which the two attributes from  Natural Language are category and position of objects that are extracted by grammar parser. In \cite{walter2013learning}, a framework for capturing spatial relationships between objects and locations, inherited from the dialog, for learning a human-centered model is presented. Works that are dealing with maps are bounded to understanding verbal position and category of objects. Also, works that aim to capture more verbal attributes are using a shallow grammar parser, inheriting from the dependency tree of the sentence; they cannot capture attributes from different linguistic expressions, where the dependency of words is not so reliable, in most realistic cases. To overcome these shortcomings, we follow our previous work \cite{faridghasemnia2019capturing}, which can capture seven different attributes from different linguistic expressions. 

Focus on attributes in knowledge representation is not on capturing attributes from different input modalities, nor grounding them to physical environment, but rather on the relationship between attributes, and reasoning over attributes. Rules are widely used as a way to express the relationship between attributes. Inductive logic programming, born at the intersection of machine learning and logic programming, is widely used as a relational learning approach \cite{lisi2012learning}. Inductive Logic programming learns rules from positive and negative examples, supported by background knowledge. The resulting rules should entail as many as positive examples, regarding background knowledge, and as few negative examples as possible \cite{muggleton1991inductive}, \cite{cropper2020turning}. In most realistic applications there is not a particular rule that includes all positive examples and avoids all negatives at the same time: to address this problem, Raedt et al. integrated probabilities with logic programming, both in deductive \cite{de2007problog} and inductive \cite{de2010probabilistic} reasoning. While these works assume an initial knowledge base, in our work we use these methods in a case where background knowledge, negative examples and positive examples are created incrementally from vision and dialog.


\section{Abstract Relational Learning Framework}
\label{sec:schema}
\begin{figure}
\centering
\includegraphics[width=85mm]{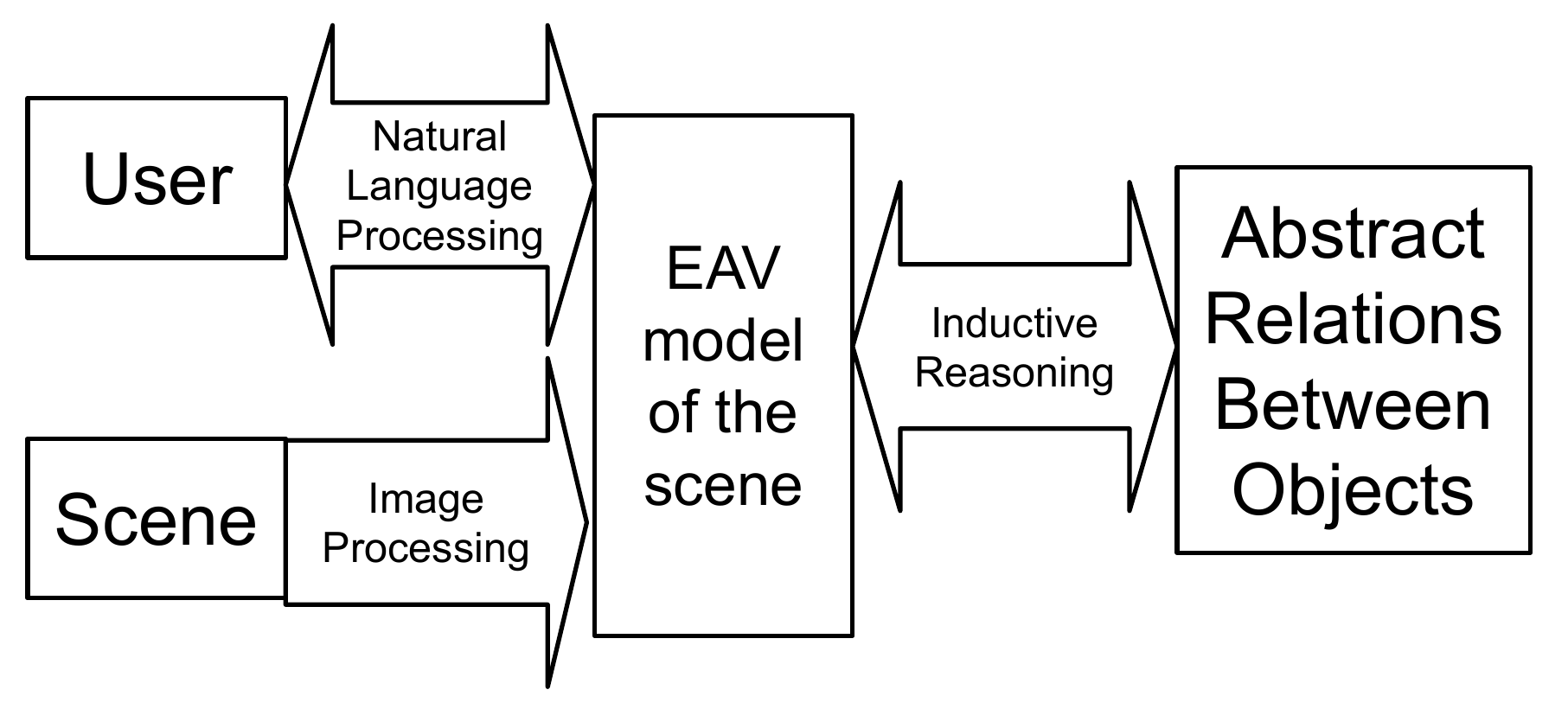}
\caption{Finding abstract relations in a scene}
\label{fig:abstractForm}
\end{figure}

In this section, we formulate the third level of understanding to be solvable by existing methodologies. As we have mentioned earlier in Introduction, the third level of understanding concerns finding the relation between entities. For a robot that acts in an environment, the third level of understanding can be interpreted as finding the latent relation between attributes of objects in the scene, specifically relations in the semantic model of the scene. We represent the semantic model of a scene by a triple store, via an Entity-Attribute-Value (EAV) model. In particular, we can model objects in the robot's surrounding environment as entities and their attributes, where each attribute represents a particular characteristic of an entity and it has a value. Once the robot has found the EAV model of a scene, it can attempt to understand the latent relationship between entities. Fig. \ref{fig:abstractForm} shows the process of finding abstract relationships in a scene, through Human-Robot Interaction. We divide this process into two problems, firstly how to find EAV model of the environment, and secondly, how to find a descriptive language that captures existing relationships in EAV model.

\subsection{EAV model of a scene}
The problem of finding EAV model of a scene can be translated into the problem of finding entities and finding their attributes. We consider entities in the scene reduced to objects in the scene, where decent systems for robot's object detection exist. The robot visual perception not only can recognize entities, but also recognize the attribute values of each detected entity. Attributes such as color, category, shape, etc. can be visually perceived via state of the art systems. Following the fruit example in the introduction, the EAV model of the scene is each individual fruit as entities and  \textit{freshness} and \textit{category} are their attributes. Creating an EAV model for this example includes creating entities, and assign values ('spoiled', 'apple', etc.) to each entity’s attributes.

There are attributes, like \textit{owner}, that are either difficult or intrinsically impossible to perceive by vision. Such attributes may occur in Human-Robot Interaction, and the robot can obtain information about their value from Natural Language. Natural Language helps a robot to enrich its EAV model of a scene via different attributes, even those that are not visually perceivable, as restrictions that might apply to an entity. 

Another source of attributes are ontologies, which can provide a deeper notion of scene semantics. In general, an ontology may contain many relevant attributes for an entity.
Although ontologies might be used in future works, in this work we found visually perceived attributes and Natural Language two sufficient sources for the EAV model of the scene.

\subsection{Inductive reasoning}
The second sub-problem is to how to find the relationship between attributes of entities. Such relationships may be presented in different languages; We chose First Order Logic (FOL) among possible languages for presentation, as it has sufficient tools for interpretability and generalizability. 

We can translate the EAV model of a scene as the robot's pieces of evidence (facts), and the relationship between facts are sets of logical formulas that we want to obtain. This computation can be done by Inductive Logic. In other words, given some facts, the role of Inductive Logic is to arrive at a formula, which should support given facts. In particular, the given facts must include a set of positive examples, and a set of negative examples, and the induced rule should include all positives and avoid all negatives, as many as possible. The inductive engine takes a specific pair of attribute and desired value, name it query, and find other attributes which entail the given query. An Inductive Logic engine can induce a logical rule to describe sufficient conditions of an entity to have a specific attribute value is to have certain attribute values.


\section{Implementation}
\label{sec:proposedFramework}

\begin{figure}
\centering
\includegraphics[width=80mm]{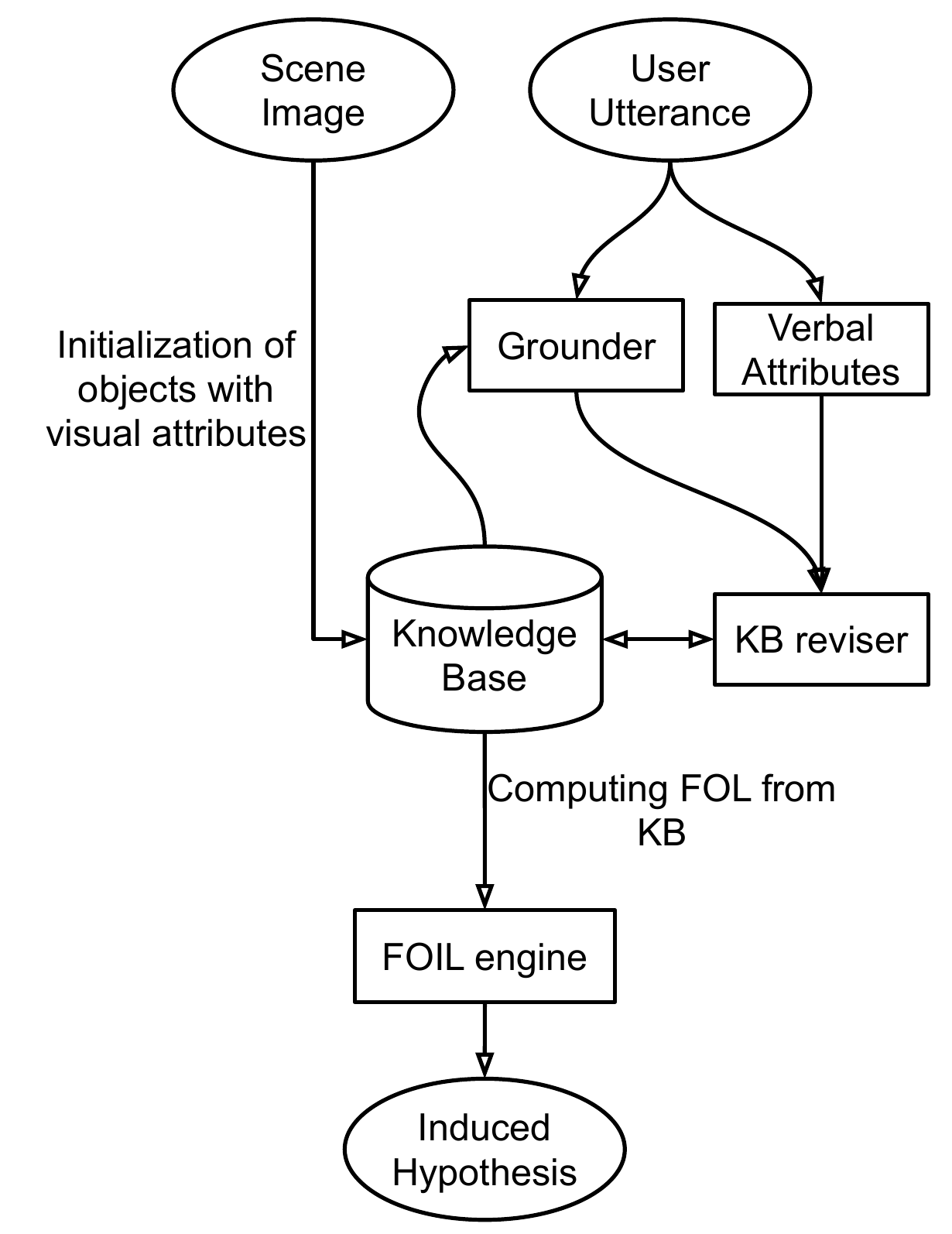}
\caption{Overview of the system.}
\label{fig:overview}
\end{figure}

In this section, we describe an implementation of our proposed framework, going from utterances and RGB camera data to logic rules. The implementation diagram of our proposed framework is depicted in Fig. \ref{fig:overview}. In short, the Grounder module, chooses one object, whose attribute the user is referring to. The KB Reviser module updates the given attribute of the selected object. The implemented Knowledge Base is a collection of all objects and their attributes in EAV format, that is shared with Grounder, KB Reviser, and the Inductive Logic module. The set of facts is obtained from the computed EAV model of the scene, and the output of FOIL (First-Order Inductive Logic) engine is induced rules. In the following sub-sections, we describe the details of each module.

\subsection{Verbal attributes}
\label{sec:language}

Computing attributes from Natural Language is a difficult problem, when different linguistic expressions have to be taken into account. We use a frame-like structure similar to FrameNet \cite{baker1998berkeley}, as a means of language formalization. Such formalization helps to find the attribute from frame type, and its value from frame elements. In this work, our framework can capture seven different attributes from language as: \textit{ownership, functionality, restriction, weight, size, label, position}.For example, in the utterance "the desk is broken", frame of this sentence is \textit{being\_operational}, and the elements 'the desk' and 'broken' are \textit{object} and \textit{lexical\_unit}, respectively. In this example, the user is assigning \textit{functionality} of the 'desk' is \textit{not\_working} by using 'broken'.

To this end, we use our deep neural model described in \cite{faridghasemnia2019capturing} for predicting frame type and its elements in an utterance. The input of this model is a sentence embedded by ConceptNet Numberbatch \cite{speer2017conceptnet}. The choice of Numberbatch is because of the embedded information of ConceptNet knowledge graph, which helps a system to be more flexible for different linguistic expressions. The output of this model is frame type and frame elements, which are formed as the attribute and the attribute value. 

\subsection{Visual attributes}
In this work we focus on two visual attributes, which are \textit{category} and \textit{color} of the objects in an image. We deployed the trained model described in \cite{Lin2017Focal}, trained on Google Open Images \cite{OpenImages} by Fizyr\footnote{https://github.com/fizyr} group. This model can predict up-to 500 different object categories, where some of these predictions might be in the same region. To overcome this, we applied a similarity threshold for predicted bounding boxes, and choose the most confident detection for each region of the image. The \textit{color} of the objects is computed based on HSV similarity between the color of the cropped region and standard colors. The name of the recognized color is chosen as the value of \textit{color} attribute.

\subsection{Grounder}
\label{sec:grounder}
We used a typical grounding methodology enhanced with semantic similarity of attributes. The framework deploys the grounding module when the user is referring to the location of an object, e.g. 'the book on the table'. This referential expression might have other attributes in addition of location, such as \textit{label}, \textit{color}, \textit{size}, and \textit{weight} of the referred object. Note that the \textit{label} is different than the \textit{category}, as the \textit{label} is a verbal attribute, and \textit{category} is a visual attribute. The problem of grounding is to find the most compatible object with respect to the referential expression attributes.

As a result of \cite{faridghasemnia2019capturing}, the neural model is trained on the single-attribute sentences cannot effectively predict multiple attributes in a sentence. On the other hand, referential grounding concerns finding multiple attributes in one given sentence. To overcome this problem and recognize multiple attributes of an object in the sentence. We use SpaCy\footnote{https://spacy.io/} dependency tree of the sentence to extract the aforementioned attributes of the sentence. Let's assume in a given sentence, \textit{n} attributes can be extracted as $dial\_attr_a$. Notice that since 
$dial\_attr_a$ are extracted from dependency tree, we do not have the type (understanding) of the attribute, and it can be seen as a symbol that describes the object.
For visual perceived objects, let's suppose there are $\{n \in N \}$ detected objects in the scene, and each object might have $a \in A$ attribute as \textit{$vis\_attr_{n,a}$}, where $n$ refers to a particular object, and $a$ belongs to the set of all attributes that has been captured for that particular object in the KB. We use the following formula to find the object in the scene most compatible with the given utterance:

\begin{equation}
\min\limits_{\forall n \in N}\{\frac{1}{|A|}\sum\limits_{\forall a \in A}distance(dial\_attr_a, vis\_attr_{n,a})\}
\end{equation}

Where the distance of $vis\_attr$ and $dial\_attr$ is the cosine distance of the corresponding symbols in Numberbatch semantic space. This simple approach lets this module ground referential expression to an entity with respect to the similarity of attributes. For example, the symbol 'mug' is grounded to the object detected with the 'cup' \textit{category}, or the symbol 'red' is easily grounded to 'purple' \textit{color} if there is no 'red' in detected colors.

In the knowledge base, objects are represented as entities and their attributes by unique ids. Once an object is selected by the grounding module, the user can assign new attributes, or update its attributes, using KB Reviser module. In other words, KB reviser module takes the selected object and verbal attributes, and update attributes of the selected object in the KB.  

\subsection{FOIL engine}
\label{sec:inductiveLogic}
A FOIL engine takes positive and negative facts and computes a general FOL formula that cover positive facts and avoids negative facts, as many as possible. The input of FOIL engine is the translated EAV model of the scene to some positive and negative facts.

Translation of the EAV model to logic formulas is not straight forward, and different approaches may be taken. We translate an EAV model into two sets of facts; The first set of facts define attribute values of entities as \textit{attributeValue(attributeID)}, where \textit{attributeValue} represents the value that each particular attribute holds, and \textit{attributeID} is a dummy argument that will be grounded in the second set of facts.

The second set of facts represents entities with respect to the given query. Suppose the goal is to find a relationship between entities that describes the condition for an entity to have a particular 'value-p' in an attribute. We form the entities in the scene toward 'value-p' as \textit{value-p(attributeId1, attribteId2,...)}, which its Natural Language translation is the condition for an entity in the scene to have a particular attribute with \textit{value-p} is the validity of \textit{attributeValue} that is linked via \textit{attributeId}s. 

 We chose probFOIL \cite{de2010probabilistic} as FOIL engine among other FOIL engines, as it has the possibility of computing fact probabilities.  Although we do not use this feature in our current experiments, we plan to leverage it in future works to account for the uncertainty in the data acquired from perception. Let us describe the process of translation of the knowledge base into probFOIL language. Suppose the scene EAV model in the KB is: \\
 {\small
 [\textit{\{objectId:obj1, category: apple, color:red, owner:harry\}, \\
 \{objectId:obj2, category: pear, color:green, owner:harry\}, \\
 \{objectId: obj3, category:pear, color:yellow, owner:hermoine\}, \\
 \{objectId: obj4, category:apple, color:yellow, owner:hermoine\}}]. 
 }

We chose to model this knowledge into entities, attributes and values, where entities in the knowledge are coded as positive and negative examples of FOIL, and attributes of entities are coded as the indices of predicates, and attribute values are coded as argument predicates. Suppose we ask the FOIL engine which objects belong to Hermoine, it should arrive to a formula which indicates that objects that are yellow belong to Hermoine. Notice that this query concerns the problem of finding sufficient conditions for the \textit{owner} attribute to take the 'hermoine' value.  To this aim, we can write the entities in the form of predicates, where the index of each predicate argument corresponds to a particular attribute, and the predicate arguments are the values corresponding to the indices of attributes. So the positive examples of entities can be written as: 

\begin{flushleft}
\textit{1.0::hermoine(cat1, col3).\\
1.0::hermoine(cat2, col3).
}
\end{flushleft}

and the negative examples of entities can be defined from objects that belong to \textit{harry} (the other owner in \textit{owner}s of entities), which becomes:

\begin{flushleft}
\textit{0.0::hermoine(cat2, col1).\\
0.0::hermoine(cat1, col2).
}
\end{flushleft}

where the argument of each entity predicate is declared in value predicates as follows:

\begin{flushleft}
\textit{red(col1).\\
green(col2).\\
yellow(col3).\\
pear(cat1).\\
apple(cat2).
}
\end{flushleft}

Notice that arguments of predicates (e.g. \textit{col1, cat1}) are meaning-less, and any arbitrary symbol can be used as long as a value is represented with a unique symbol in all predicates (e.g. all pears represented by \textit{cat1}). 

The setting of probFOIL contains the target, type, and modes, where the target is obtained from the query (e.g. \textit{hermoine/2}, which is predicate/arity), type of arguments are the name of attributes, and by mode, we forced probFoil so that each argument should already exist when a literal is added \cite{de2010probabilistic}.

\section{Demonstration and Discussion}
\label{sec:demo}
In this section, we demonstrate an interaction between the user and the system and discuss different applications of the proposed system.
\subsection{Showcased example}
\begin{figure}
\centering
\includegraphics[width=0.49\textwidth]{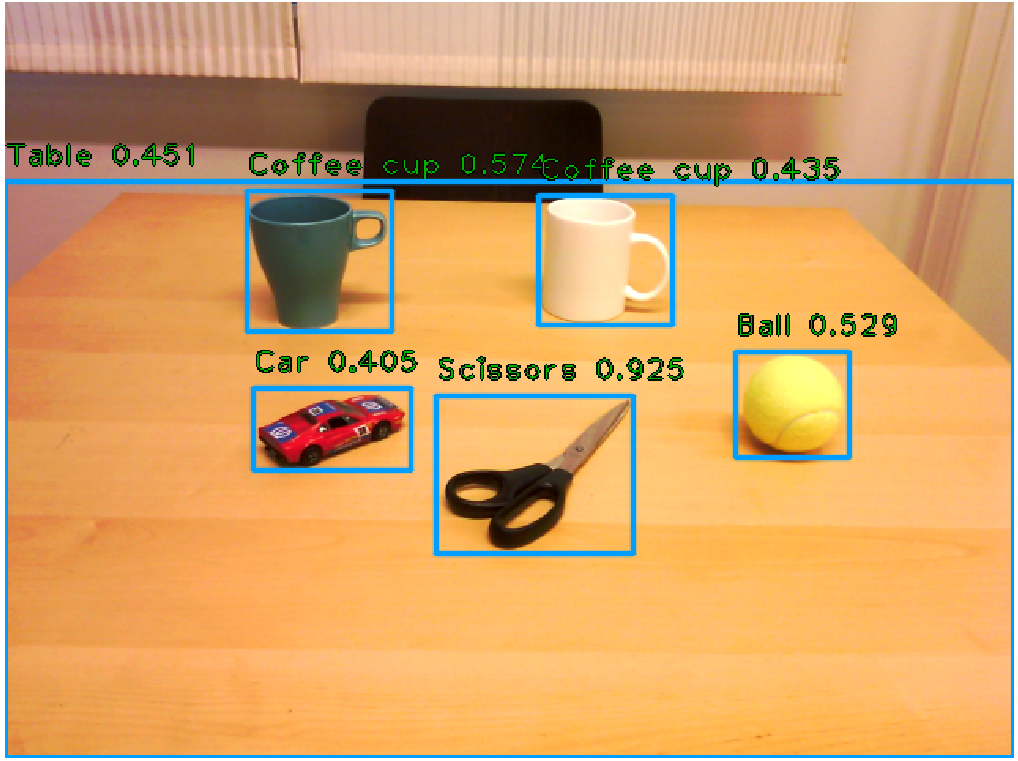}
\caption{Scene from Pepper view.}
\label{fig:scenePep}
\end{figure}

\begin{figure}
\centering
\includegraphics[height=0.49\textwidth, angle=90]{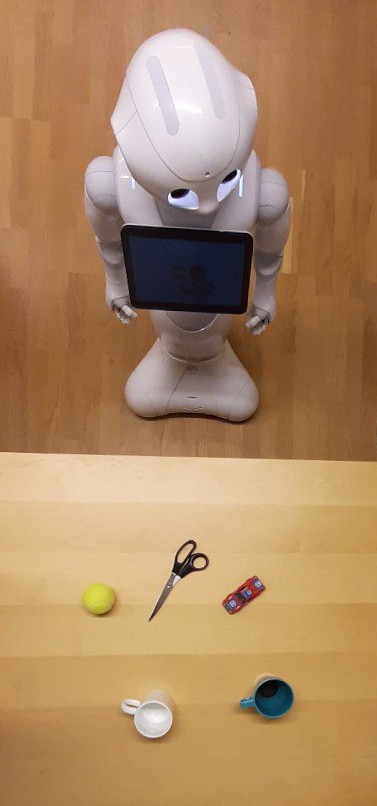}
\caption{Experimental setup (Top view).}
\label{fig:scenetopView}
\end{figure}

In this section, we describe a simple scenario which let us discuss the feasibility of our proposal. Consider a scenario where the user is describing the scene shown in Fig. \ref{fig:scenetopView}. The pepper visual perception of the scene is shown in Fig. \ref{fig:scenePep}. The user is giving attributes through verbal interactions that is transcribed below. Note that different attribute relations can be figured out by our system, where in this scenario we just showed a simplified example that four out of 5 detected objects are enriched by four attributes, while the blue cup has only two attributes captured through visual perception. 

\begin{flushright}
\vspace{-2pt}
\textit{\textbf{Human}}
\vspace{2pt}
\tcbox[enhanced,size=fbox,drop shadow southwest,
    sharp corners]{Hello.}
\end{flushright}  

\begin{flushleft}
\vspace{-40pt}
\textit{\textbf{Robot}}
\vspace{18pt}
\tcbox[enhanced,size=fbox,drop shadow southeast,
    sharp corners]{Hi there!}
\end{flushleft}

\begin{flushright}
\vspace{-10pt}
\tcbox[enhanced,size=fbox,drop shadow southwest,
    sharp corners]{the white mug on the table}
\end{flushright}  

\begin{flushleft}
\vspace{-10pt}
\tcbox[enhanced,size=fbox,drop shadow southeast,
    sharp corners]{I see, the white mug is on the table}
\end{flushleft}

\begin{flushright}
\vspace{-10pt}
\tcbox[enhanced,size=fbox,drop shadow southwest,
    sharp corners]{i guess it is for mary}
\end{flushright}  

\begin{flushleft}
\vspace{-10pt}
\tcbox[enhanced,size=fbox,drop shadow southeast,
    sharp corners]{Got it! it belongs to mary}
\end{flushleft}

\begin{flushright}
\vspace{-10pt}
\tcbox[enhanced,size=fbox,drop shadow southwest,
    sharp corners]{its label is kitchenware}
\end{flushright}  

\begin{flushleft}
\vspace{-10pt}
\tcbox[enhanced,size=fbox,drop shadow southeast,
    sharp corners]{Ok, I save it as kitchenware}
\end{flushleft}

\begin{flushright}
\vspace{-10pt}
\tcbox[enhanced,size=fbox,drop shadow southwest,
    sharp corners]{the scissor on the table}
\end{flushright}  

\begin{flushleft}
\vspace{-10pt}
\tcbox[enhanced,size=fbox,drop shadow southeast,
    sharp corners]{I see, the scissor is on the table}
\end{flushleft}

\begin{flushright}
\vspace{-10pt}
\tcbox[enhanced,size=fbox,drop shadow southwest,
    sharp corners]{also it is for mary}
\end{flushright}  

\begin{flushleft}
\vspace{-10pt}
\tcbox[enhanced,size=fbox,drop shadow southeast,
    sharp corners]{Got it! it belongs to mary}
\end{flushleft}

\begin{flushright}
\vspace{-10pt}
\tcbox[enhanced,size=fbox,drop shadow southwest,
    sharp corners]{name it kitchenware}
\end{flushright}  

\begin{flushleft}
\vspace{-10pt}
\tcbox[enhanced,size=fbox,drop shadow southeast,
    sharp corners]{Ok, I save it as kitchenware}
\end{flushleft}

\begin{flushright}
\vspace{-10pt}
\tcbox[enhanced,size=fbox,drop shadow southwest,
    sharp corners]{the tennis ball on the table}
\end{flushright}  

\begin{flushleft}
\vspace{-10pt}
\tcbox[enhanced,size=fbox,drop shadow southeast,
    sharp corners]{I see, the tennis ball is on the table}
\end{flushleft}

\begin{flushright}
\vspace{-10pt}
\tcbox[enhanced,size=fbox,drop shadow southwest,
    sharp corners]{the label is toy}
\end{flushright}  

\begin{flushleft}
\vspace{-10pt}
\tcbox[enhanced,size=fbox,drop shadow southeast,
    sharp corners]{Ok, I save it as toy}
\end{flushleft}

\begin{flushright}
\vspace{-10pt}
\tcbox[enhanced,size=fbox,drop shadow southwest,
    sharp corners]{it belongs to toby}
\end{flushright}  

\begin{flushleft}
\vspace{-10pt}
\tcbox[enhanced,size=fbox,drop shadow southeast,
    sharp corners]{Got it! it belongs to toby}
\end{flushleft}

\begin{flushright}
\vspace{-10pt}
\tcbox[enhanced,size=fbox,drop shadow southwest,
    sharp corners]{the car on the table}
\end{flushright}  

\begin{flushleft}
\vspace{-10pt}
\tcbox[enhanced,size=fbox,drop shadow southeast,
    sharp corners]{I see, the car is on the table}
\end{flushleft}

\begin{flushright}
\vspace{-10pt}
\tcbox[enhanced,size=fbox,drop shadow southwest,
    sharp corners]{it also belongs to toby}
\end{flushright}  

\begin{flushleft}
\vspace{-10pt}
\tcbox[enhanced,size=fbox,drop shadow southeast,
    sharp corners]{Got it! it also belongs to toby}
\end{flushleft}

\begin{flushright}
\vspace{-10pt}
\tcbox[enhanced,size=fbox,drop shadow southwest,
    sharp corners]{save the label is toy}
\end{flushright}  

\begin{flushleft}
\vspace{-10pt}
\tcbox[enhanced,size=fbox,drop shadow southeast,
    sharp corners]{Ok, I save it as toy}
\end{flushleft}

A user can insert information about the objects in the scene, through verbal interaction, and at each turn the proposed system utters a template-based acknowledgment utterance. Following the interactions, we can ask the system to find relationships between different attributes. As an example, if we ask the system which objects belong to 'mary', the system computes \textit{mary(A,B,C,D) :- kitchenware(C)}. This formula shows that the sufficient condition for an object to belong to 'mary' is that it should have 'kitchenware' \textit{label}. Also if we query which objects has 'toy' \textit{label}, it gives \textit{toy(A,B,C,D) :- toby(C)}. This formula declares that any object that belongs to 'toby' has 'toy' \textit{label}. Notice that in these formulas, arguments \textit{A, B, C, D} represent different attributes that the system captured through interaction, which are \textit{category}, \textit{color}, \textit{label}, \textit{owner} and \textit{location}. In the following sub-section, we discuss different applications of our proposed system.

\subsection{Discussion}
\label{sec:discussion}
Finding relations between attributes can be used for auto-enriching attributes of the objects in a scene, or use the computed formula to refine captured attributes. For example, if we ask the system in this scene which objects are on the table, since the \textit{location} of all objects are 'on\_table', it will give the formula \textit{on\_table :- true}, which means all the objects are on the table; In other words, the location of the blue mug can be inferred. Moreover, suppose the user assigns 'kitchenware' to the blue mug; As the obtained rule indicates that all 'kitchenware's belongs to 'mary', so the robot can assume that also the blue mug belongs to 'mary'. 

One of the interesting applications of this framework is resolving ambiguities in robot's sensory input; In particular, the \textit{category} of detected objects. Neural Networks can detect objects in any region of the image, but detection is not always reliable. Notice that usually, Neural Networks provide multiple detected classes for a region in the scene, and the standard approach is to choose the detection with the highest confidence and ignore the rest. This scheme can be improved when additional information about common attributes is available. In other words, a logical formula between attributes of objects can be used to find which detection is the most probable one, even it might not have the highest confidence of the Neural Network. For example, consider a scene with four different white mugs, while three of them are detected as mugs, and the other detected object is 80 \% bowl and 50\% mug. In this scene, a logical rule (e.g. \textit{mug(A):- white(A).}) can show that all objects that are white are mugs; Such a rule can help the robot to choose the proper \textit{category} with lower confidence. Although this rule cannot be generalized to all white objects in the robot's knowledge are white, but with the proper configuration of context and a rich representation of entities, this rule can be reliable. For example if the mug is situated in the cabinet of mugs, and entity attributes are \textit{shape}: 'cylindrical with handle', \textit{color}:'white', \textit{position}:'mug's cabinet', this logical rule can be used in resolving ambiguity in robot's scene understanding. 

The number of attributes plays a key role in this framework. In particular, with a higher number of visually perceived attributes, the inductive engine can unravel the relationships between entity attributes with less human effort. In our setting, the perceived attributes are bound to color, and \textit{category} of objects, while a richer visual object annotator can minimize the verbal interactions. The low number of attributes becomes more conspicuous when there is high diversity in attribute values. Consider the example of four mugs in a scene, where each mug has different color. In this situation the framework cannot figure out any rule between \textit{category} and \textit{color} of objects.

\section{Conclusion}
\label{conclusion}
In this work, we proposed a framework to compute First-Order Logical formulas that represent latent relations between entities in a scene semantic model. Our proposed framework creates a semantic model of the robot's  environment, in the format of entities and attributes, from visual perception and dialogic interaction with the user. This semantic model is further used for finding the latent relations between entities in the form of First-Order Logic. The obtained rules specify which attributes entail a specific property in an entity, and can be used by the robot for removing uncertainty in sensory input, or to enrich its knowledge base.

\bibliographystyle{plain} 
\bibliography{bibliography} 

\begin{thebibliography}{10}

\bibitem{baker1998berkeley}
Collin~F Baker, Charles~J Fillmore, and John~B Lowe.
\newblock The berkeley framenet project.
\newblock In {\em Proceedings of the 17th international conference on
  Computational linguistics-Volume 1}, pages 86--90. Association for
  Computational Linguistics, 1998.

\bibitem{coradeschi2003introduction}
Silvia Coradeschi and Alessandro Saffiotti.
\newblock An introduction to the anchoring problem.
\newblock {\em Robotics and autonomous systems}, 43(2-3):85--96, 2003.

\bibitem{cropper2020turning}
Andrew Cropper, Sebastijan Duman{\v{c}}i{\'c}, and Stephen~H Muggleton.
\newblock Turning 30: New ideas in inductive logic programming.
\newblock {\em arXiv preprint arXiv:2002.11002}, 2020.

\bibitem{de2007problog}
Luc De~Raedt, Angelika Kimmig, and Hannu Toivonen.
\newblock Problog: A probabilistic prolog and its application in link
  discovery.
\newblock In {\em IJCAI}, volume~7, pages 2462--2467. Hyderabad, 2007.

\bibitem{de2010probabilistic}
Luc De~Raedt and Ingo Thon.
\newblock Probabilistic rule learning.
\newblock In {\em International conference on inductive logic programming},
  pages 47--58. Springer, 2010.

\bibitem{DongLanguage2016}
Li~Dong and Mirella Lapata.
\newblock Language to logical form with neural attention.
\newblock {\em Proceedings of the 54th Annual Meeting of the Association for
  Computational Linguistics (Volume 1: Long Papers)}, 2016.

\bibitem{Farhadi2009Describing}
Ali Farhadi, Ian Endres, Derek Hoiem, and David~A. Forsyth.
\newblock Describing objects by their attributes.
\newblock {\em 2009 IEEE Conference on Computer Vision and Pattern
  Recognition}, pages 1778--1785, 2009.

\bibitem{faridghasemnia2019capturing}
Mohamadreza Faridghasemnia, Andrea Vanzo, and Daniele Nardi.
\newblock Capturing frame-like object descriptors in human augmented mapping.
\newblock In {\em International Conference of the Italian Association for
  Artificial Intelligence}, pages 392--404. Springer, 2019.

\bibitem{harnad1990symbol}
Stevan Harnad.
\newblock The symbol grounding problem.
\newblock {\em Physica D: Nonlinear Phenomena}, 42(1-3):335--346, 1990.

\bibitem{Isola2015DiscoveringSA}
Phillip Isola, Joseph~J. Lim, and Edward~H. Adelson.
\newblock Discovering states and transformations in image collections.
\newblock {\em 2015 IEEE Conference on Computer Vision and Pattern Recognition
  (CVPR)}, pages 1383--1391, 2015.

\bibitem{Kazemzadeh2014Refer}
Sahar Kazemzadeh, Vicente Ordonez, Mark Matten, and Tamara Berg.
\newblock Referitgame: Referring to objects in photographs of natural scenes.
\newblock pages 787--798, 01 2014.

\bibitem{krishnavisualgenome}
Ranjay Krishna, Yuke Zhu, Oliver Groth, Justin Johnson, Kenji Hata, Joshua
  Kravitz, Stephanie Chen, Yannis Kalantidis, Li-Jia Li, David~A Shamma,
  Michael Bernstein, and Li~Fei-Fei.
\newblock Visual genome: Connecting language and vision using crowdsourced
  dense image annotations.
\newblock 2016.

\bibitem{kushman2013using}
Nate Kushman and Regina Barzilay.
\newblock Using semantic unification to generate regular expressions from
  natural language.
\newblock In {\em Proceedings of the 2013 Conference of the North {A}merican
  Chapter of the Association for Computational Linguistics: Human Language
  Technologies}, pages 826--836, Atlanta, Georgia, June 2013. Association for
  Computational Linguistics.

\bibitem{OpenImages}
Alina Kuznetsova, Hassan Rom, Neil Alldrin, Jasper Uijlings, Ivan Krasin, Jordi
  Pont-Tuset, Shahab Kamali, Stefan Popov, Matteo Malloci, Alexander
  Kolesnikov, Tom Duerig, and Vittorio Ferrari.
\newblock The open images dataset v4: Unified image classification, object
  detection, and visual relationship detection at scale.
\newblock {\em IJCV}, 2020.

\bibitem{liang2016learning}
Percy Liang.
\newblock Learning executable semantic parsers for natural language
  understanding.
\newblock {\em Communications of the ACM}, 59(9):68--76, 2016.

\bibitem{Lin2017Focal}
Tsung-Yi Lin, Priya Goyal, Ross Girshick, Kaiming He, and Piotr Dollar.
\newblock Focal loss for dense object detection.
\newblock {\em 2017 IEEE International Conference on Computer Vision (ICCV)},
  Oct 2017.

\bibitem{lisi2012learning}
Francesca~A. Lisi.
\newblock Learning onto-relational rules with inductive logic programming,
  2012.

\bibitem{Mao2016Generation}
Junhua Mao, Jonathan Huang, Alexander Toshev, Oana Camburu, Alan Yuille, and
  Kevin Murphy.
\newblock Generation and comprehension of unambiguous object descriptions.
\newblock {\em 2016 IEEE Conference on Computer Vision and Pattern Recognition
  (CVPR)}, Jun 2016.

\bibitem{muggleton1991inductive}
Stephen Muggleton.
\newblock Inductive logic programming.
\newblock {\em New generation computing}, 8(4):295--318, 1991.

\bibitem{pronobis2012large}
Andrzej Pronobis and Patric Jensfelt.
\newblock Large-scale semantic mapping and reasoning with heterogeneous
  modalities.
\newblock In {\em Robotics and Automation (ICRA), 2012 IEEE International
  Conference on}, pages 3515--3522. IEEE, 2012.

\bibitem{OlgaAttribute2012}
Olga Russakovsky and Li~Fei-Fei.
\newblock Attribute learning in large-scale datasets.
\newblock In Kiriakos~N. Kutulakos, editor, {\em Trends and Topics in Computer
  Vision}, pages 1--14, Berlin, Heidelberg, 2012. Springer Berlin Heidelberg.

\bibitem{Sadeghi2011Recognition}
M.~A. {Sadeghi} and A.~{Farhadi}.
\newblock Recognition using visual phrases.
\newblock In {\em CVPR 2011}, pages 1745--1752, June 2011.

\bibitem{speer2017conceptnet}
Robyn Speer, Joshua Chin, and Catherine Havasi.
\newblock {ConceptNet} 5.5: An open multilingual graph of general knowledge.
\newblock pages 4444--4451, 2017.

\bibitem{steedman2011combinatory}
Mark Steedman and Jason Baldridge.
\newblock Combinatory categorial grammar.
\newblock {\em Non-Transformational Syntax: Formal and explicit models of
  grammar}, pages 181--224, 2011.

\bibitem{vanzo2018incrementally}
Andrea Vanzo, Jose~L Part, Yanchao Yu, Daniele Nardi, and Oliver Lemon.
\newblock Incrementally learning semantic attributes through dialogue
  interaction.
\newblock In {\em Proceedings of the 17th International Conference on
  Autonomous Agents and MultiAgent Systems}, pages 865--873. International
  Foundation for Autonomous Agents and Multiagent Systems, 2018.

\bibitem{walter2013learning}
Matthew~R Walter, Sachithra Hemachandra, Bianca Homberg, Stefanie Tellex, and
  Seth Teller.
\newblock Learning semantic maps from natural language descriptions.
\newblock Robotics: Science and Systems, 2013.

\bibitem{Yu2018Mattnet}
Licheng Yu, Zhe Lin, Xiaohui Shen, Jimei Yang, Xin Lu, Mohit Bansal, and
  Tamara~L. Berg.
\newblock Mattnet: Modular attention network for referring expression
  comprehension.
\newblock {\em 2018 IEEE/CVF Conference on Computer Vision and Pattern
  Recognition}, Jun 2018.

\end{thebibliography}

\end{document}